\documentclass[nohyperref]{article}

\usepackage{microtype}
\usepackage{graphicx}
\usepackage{subfigure}
\usepackage{booktabs} %
\usepackage{todonotes} 

\usepackage{hyperref}
\ifdefined\nohyperref\else\ifdefined\hypersetup
  \definecolor{mydarkblue}{rgb}{0,0.50,0.30}
  \hypersetup{ %
    pdftitle={},
    pdfauthor={},
    pdfsubject={},
    pdfkeywords={},
    pdfborder=0 0 0,
    pdfpagemode=UseNone,
    colorlinks=true,
    linkcolor=mydarkblue,
    citecolor=mydarkblue,
    filecolor=mydarkblue,
    urlcolor=mydarkblue,
    pdfview=FitH}

  \ifdefined\isaccepted \else
    \hypersetup{pdfauthor={Anonymous Submission}}
  \fi
\fi\fi

\usepackage[utf8]{inputenc}
\usepackage{xcolor}
\usepackage{bbold}
\usepackage{enumitem}
\usepackage{hyperref}

\usepackage{bm}
\usepackage{amssymb,amsmath}
\usepackage{amsthm}
\usepackage{mathtools}
\mathtoolsset{showonlyrefs}

\usepackage{booktabs}
\usepackage{wrapfig}

\usepackage[accepted]{icml2023}

\usepackage{mdframed}
\definecolor{theoremcolor}{rgb}{255, 255, 255}
\mdfsetup{
    backgroundcolor=theoremcolor,
    linewidth=0.5pt,
}

\usepackage{algorithm} 

\newmdtheoremenv{definition}{Definition}
\newmdtheoremenv{proposition}{Proposition}
\newmdtheoremenv{corollary}{Corollary}
\newmdtheoremenv{theorem}{Theorem}
\newmdtheoremenv{algo}{Algorithm}
\newmdtheoremenv{lemma}{Lemma}

\icmltitlerunning{Fast, Differentiable and Sparse Top-k: a Convex Analysis Perspective}

\DeclareMathOperator*{\argmax}{argmax}
\DeclareMathOperator*{\argmin}{argmin}

\DeclareMathOperator*{\conv}{conv}

\def\RR{\mathbb{R}}

\def\a{\bm{a}}
\def\b{\bm{b}}
\def\t{\bm{t}}
\def\p{\bm{p}}
\def\q{\bm{q}}
\def\s{\bm{s}}
\def\u{\bm{u}}
\def\v{\bm{v}}
\def\w{\bm{w}}
\def\x{\bm{x}}
\def\y{\bm{y}}
\def\z{\bm{z}}
\def\alphav{\bm{\alpha}}
\def\thetav{\bm{\theta}}
\def\sigmav{\bm{\sigma}}
\def\rhov{\bm{\rho}}
\def\ones{\mathbf{1}}
\def\zeros{\mathbf{0}}

\def\argsort{\mathrm{argsort}}
\def\sort{\mathrm{sort}}
\def\rank{\mathrm{rank}}
\def\topk{\mathrm{topk}}
\def\topkmag{\mathrm{topkmag}}
\def\topkmask{\mathrm{topkmask}}

\begin{document}

\twocolumn[
\icmltitle{Fast, Differentiable and Sparse Top-k: a Convex Analysis Perspective}

\icmlsetsymbol{equal}{*}

\begin{icmlauthorlist}
\icmlauthor{Michael E. Sander}{equal,ens}
\icmlauthor{Joan Puigcerver}{google}
\icmlauthor{Josip Djolonga}{google}
\icmlauthor{Gabriel Peyré}{ens,cnrs}
\icmlauthor{Mathieu Blondel}{google}
\end{icmlauthorlist}

\icmlaffiliation{ens}{Ecole Normale Supérieure}
\icmlaffiliation{google}{Google Research, Brain team}
\icmlaffiliation{cnrs}{CNRS}

\icmlcorrespondingauthor{Michael E. Sander}{michael.sander@ens.fr}
\icmlcorrespondingauthor{Joan Puigcerver}{jpuigcerver@google.com}
\icmlcorrespondingauthor{Josip Djolonga}{josipd@google.com}
\icmlcorrespondingauthor{Gabriel Peyré}{gabriel.peyre@ens.fr}
\icmlcorrespondingauthor{Mathieu Blondel}{mblondel@google.com}

\icmlkeywords{top-k, permutahedron, isotonic optimization, smoothing}

\vskip 0.3in
]

\printAffiliationsAndNotice{\icmlEqualContribution} %

\begin{abstract}
The top-$k$ operator returns a sparse vector, where the non-zero values correspond to the $k$ largest values of the input.  
Unfortunately, because it is a discontinuous function, it is difficult to incorporate in neural networks trained end-to-end with backpropagation.
Recent works have considered differentiable relaxations, based either on regularization or perturbation techniques.
However, to date, no approach is fully differentiable \emph{and} sparse.
In this paper, we propose new differentiable \emph{and} sparse top-$k$ operators. 
We view the top-$k$ operator as a linear program over the permutahedron, the convex hull of permutations. 
We then introduce a $p$-norm regularization term to smooth out the operator,
and show that its computation can be reduced to isotonic optimization.
Our framework is significantly more general than the existing one and allows for example to express top-$k$ operators that select values \emph{in magnitude}. 
On the algorithmic side, in addition to pool adjacent violator (PAV) algorithms,
we propose a new GPU/TPU-friendly Dykstra algorithm to solve isotonic optimization problems.   
We successfully use our operators to prune weights in neural networks,
to fine-tune vision transformers,
and as a router in sparse mixture of experts.
\end{abstract}

\section{Introduction}

Finding the top-$k$ values and their corresponding indices in a vector is a widely used building block in modern neural networks. For instance, in sparse mixture of experts (MoEs) \citep{shazeer2017outrageously, fedus2022review}, a top-$k$ router maps each token to a selection of $k$ experts (or each expert to a selection of $k$ tokens). In beam search for sequence decoding \citep{wiseman2016sequence}, a beam of $k$ possible output sequences is maintained and updated at each decoding step. 
For pruning neural networks, the top-$k$ operator can be used to sparsify a neural network, by removing weights with the smallest magnitude \citep{han2015learning, frankle2018lottery}. Finally, top-$k$ accuracy (e.g., top-$3$ or top-$5$) is frequently used to evaluate the performance of neural networks at inference time.
\begin{figure}[H]
\centering
\includegraphics[width=0.85\columnwidth]{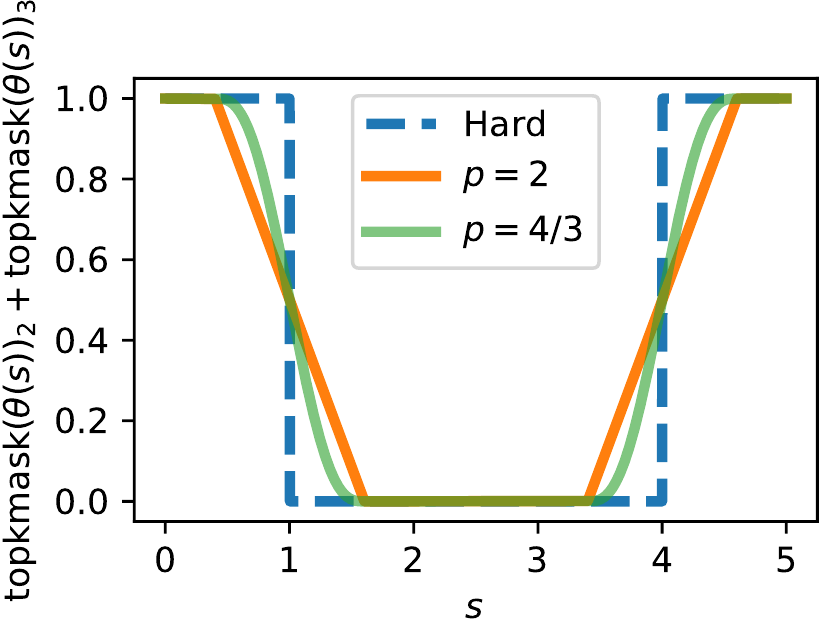} 
\caption{\textbf{Illustration of our differentiable and sparse top-k mask.} For $k=2$, we consider $\thetav(s) = (3, 1, -1 + s, s) \in \RR^4$ and plot $\mathrm{topkmask}(\thetav(s))_2 + \mathrm{topkmask}(\thetav(s))_3$ as a function of $s$. We compare the hard version (no regularization) with our proposed operator using $p$-norm regularization: $p=2$ leads to differentiable a.e. operator; $p=4/3$, leads to a differentiable operator. Both operators are \textbf{sparse}: they are exactly 0 for some values of $s$.}\label{fig:intro}
\end{figure}
\vspace{-1em}
However, the top-$k$ operator is a discontinuous piecewise affine function with derivatives either undefined or constant (the related top-$k$ \emph{mask} operator, which returns a binary encoding of the indices corresponding to the top-$k$ values, has null derivatives). This makes it hard to use in a neural network trained with gradient backpropagation. Recent works have considered differentiable relaxations, based either on regularization or perturbation techniques (see \S\ref{sec:related_work} for a review).
However, to date, no approach is differentiable everywhere \emph{and} sparse.
Sparsity is crucial in neural networks
that require \emph{conditional} computation. This is for instance the case in sparse mixture of experts, 
where the top-$k$ operator is used to ``route'' tokens to selected experts. Without sparsity,
all experts would need to process all tokens, leading to high computational cost.
This is also the case when selecting weights with highest magnitude in neural networks: the non-selected weights should be exactly $0$ in order to optimize the computational and memory costs.

In this work, we propose novel differentiable everywhere \emph{and} sparse top-$k$ operators (see Figure \ref{fig:intro}). We build upon the framework of \citet{blondel2020fast}, which casts sorting and ranking as linear programs
over the permutahedron, and uses a reduction to isotonic optimization.
We significantly generalize that framework in several ways. Specifically, we make the following contributions: %
\begin{itemize}[topsep=0pt,itemsep=2pt,parsep=2pt,leftmargin=10pt]

\item After reviewing related work in \S\ref{sec:related_work}
and background in \S\ref{sec:background}, we introduce our generalized framework
in \S\ref{sec:proposed_framework}. We introduce a new nonlinearity $\varphi$,
allowing us to express new operators (such as the top-$k$ in magnitude).
In doing so, we also establish new connections between so-called $k$-support norms and the permutahedron.

\item We introduce a regularization term to obtain a relaxed top-$k$ operator. In particular, using $p$-norm regularization, we obtain the first differentiable everywhere \emph{and} sparse top-$k$ operator (Figure \ref{fig:topk}).

\item In \S\ref{sec:algorithms}, we derive pool adjacent violator (PAV) algorithms for solving isotonic optimization when using $\varphi$ and/or when using $p$-norm regularization. We show that the Jacobian of our operator can be computed in closed form.

\item As a GPU/TPU friendly alternative to PAV, we propose a Dykstra algorithm to solve isotonic optimization, which is easy to vectorize in the case $p=2$. 

\item In \S\ref{sec:experiments}, we chose to focus on three applications of our operators. First, we use them to prune weights in a multilayer perceptron during training and show that they lead to better accuracy than with a hard top-$k$. Second, we define top-$k$ losses to fine-tune vision transformers (ViTs) and obtain better top-$k$ accuracy than with the cross-entropy loss. Finally, we use our operators as a router in vision mixture of experts, and show that they outperform the hard top-$k$ router.
\end{itemize}

\section{Related work}
\label{sec:related_work}

\paragraph{Differentiable loss functions.} 

Several works have proposed a differentiable loss function as a surrogate for a discrete, discontinuous metric. For example, loss functions have been proposed for top-$k$ accuracy \citep{lapin_2015,lapin_2016,berrada2018smooth,petersen2022differentiable} and various ranking metrics \citep{chapelle_2010,sinkprop, rolinek_2020}.

\paragraph{Differentiable operators.} 

In a different line of research, which is the main focus of this work, a differentiable operator is proposed,  which can be used either as an intermediate layer in a neural network or as final output, fed into an arbitrary loss function.
For example, \citet{niepert2021implicit} proposed a framework for computing gradients of discrete
probability distributions or optimization problems. \citet{amos_2019} and \citet{qian_2022} proposed a smooth (but not sparse) top-$k$ operator based on binary entropy regularization. Related projections on the capped simplex were proposed \citep{martins_2017,malaviya_2018,blondel_projection_2019} in different contexts. These works use ad-hoc algorithms, while we use a reduction to isotonic optimization. 
\citet{cuturi2019differentiable} proposed a relaxation of the sorting and ranking operators based on entropy-regularized optimal transport and used it to obtain a differentiable (but again not sparse) top-$k$ operator. Its computation relies on Sinkhorn's algorithm \citep{sinkhorn1967diagonal, cuturi2013sinkhorn}, which in addition makes it potentially slow to compute and differentiate. A similar approach was proposed by \citet{xie2020differentiable}. \citet{petersen2021differentiable} propose
smooth differentiable sorting networks by combining differentiable sorting functions with sorting networks. Other relaxation of the sort and operators have been proposed by \citet{groverstochastic} and  \citet{prillo2020softsort}.

The closest work to ours is that of \citet{blondel2020fast}, in which
sorting and ranking are cast as linear programs over the permutahedron. To make these operators differentiable, regularization is introduced in the formulation and it is shown that the resulting operators can be computed via isotonic optimization in $O(n \log n)$ time. 
Unfortunately, the proposed operators still include kinks: they are not differentiable everywhere. The question of how to construct
a differentiable everywhere \emph{and} sparse relaxation with $O(n \log n)$ time complexity is therefore still open.
In this work, we manage to do so by using $p$-norm regularization.
Furthermore, by introducing a new nonlinearity $\varphi$,
we significantly generalize the framework of \citet{blondel2020fast}, allowing us for instance to express a new top-$k$ operator \emph{in magnitude}.
We introduce a new GPU/TPU friendly Dykstra algorithm as an alternative to PAV.

Instead of introducing regularization, another technique relies on perturbation \citep{berthet2020learning}. 
This technique has been used to obtain a differentiable (but still \emph{not sparse}) top-k for image patch selection \citep{cordonnier2021differentiable}. 

\paragraph{Pruning weights with small magnitude.} 

Many recent works focus on neural network pruning, where parameters are removed to significantly reduce the size of a model. See \citet{blalock2020state} for a recent survey. A simple yet popular method for pruning neural networks is by global magnitude pruning \cite{collins2014memory, han2015learning}: weights with lowest absolute value are set to $0$. While most of the pruning techniques are performed after the model is fully trained \cite{blalock2020state}, some works prune periodically during training  \cite{gale2019state}. However, to the best of our knowledge, pruning by magnitude is not done in a differentiable fashion. In this work, we empirically show that pruning weights with a differentiable (or differentiable almost everywhere) top-$k$ operator in magnitude during training leads to faster convergence and better accuracy than with a ``hard'' one.

\paragraph{Top-k operator for mixture of experts.}

Sparse mixture of experts models (MoEs) \cite{shazeer2017outrageously} are a class of deep learning models where only a small proportion of the model, known as experts, is activated, depending on its input. Therefore, sparse MoEs are able to increase the number of parameters without increasing the time complexity of the model.
Sparse MoEs have achieved great
empirical successes in computer vision \cite{riquelme2021scaling, zhou2022mixture} as well as natural language processing \cite{shazeer2017outrageously, lewis2021base, fedus2021switch}. At the heart of the sparse MoE model is its routing mechanism, which determines which inputs (or tokens) are assigned to which experts. 
In the sparse mixture of experts literature, some works have recently proposed new top-$k$ operators in the routing module.
\citet{hazimeh2021dselect} proposed a binary encoding formulation to select non-zero weights. However, their formulation does not approximate the true top-$k$ operator and sparsity is only supported at inference time, not during training.
\citet{sparse_ot} proposed an optimal transport formulation supporting $k$-sparsity constraints and used it for sparse mixture of experts. In this work, we propose to replace the hard top-$k$ router, which is a discontinuous function, by our smooth relaxation.

\section{Background}
\label{sec:background}

\paragraph{Notation.}

We denote a permutation of $[n]$ by $\sigma = (\sigma_1, \dots, \sigma_n)$ and its inverse by $\sigma^{-1}$.
When seen as a vector, we denote it $\sigmav$.
We denote the set of all $n!$ permutations by $\Sigma$. 
Given a vector $\x \in \RR^n$, we denote the version of $\x$ permuted according to $\sigma$
by $\x_\sigma \coloneqq (x_{\sigma_1}, \dots, x_{\sigma_n})$.
We denote the $i$-th largest value of $\x \in \RR^n$ by $x_{[i]}$.
Without loss of generality, we always sort values in 
descending order. The conjugate of $f(\x)$ is denoted by
$f^*(\y) \coloneqq \sup_{\x \in \RR^n} \langle \x, \y \rangle - f(\x)$.

\paragraph{Review of operators.}

We denote the \textbf{argsort} operator as the permutation $\sigmav$ sorting $\x \in \RR^n$, i.e.,
\begin{equation}
\argsort(\x) \coloneqq \sigmav,
\quad \text{where} \quad
x_{\sigma_1} \ge \dots \ge x_{\sigma_n}.
\end{equation}
We denote the \textbf{sort} operator as the values of $\x \in \RR^n$ in sorted order, i.e.,
\begin{equation}
\sort(\x) \coloneqq \x_\sigma,
\quad \text{where} \quad
\sigmav = \argsort(\x).
\end{equation}
The value $[\sort(\x)]_i$ is also known as the $i$-th order statistic.
We denote the \textbf{rank} operator as the function returning the positions of the vector 
$\x \in \RR^n$ in the sorted vector. 
It is formally equal to the argsort's inverse permutation:
\begin{equation}
\rank(\x) \coloneqq \sigmav^{-1},
\quad \text{where} \quad
\sigmav = \argsort(\x).
\end{equation}
Smaller rank $[\rank(\x)]_i$ means that $x_i$ has higher value.
The \textbf{top-k mask} operator returns a bit-vector encoding whether each value $x_i$ is within the top-$k$ values or not:
\begin{equation}
[\topkmask(\x)]_i \coloneqq    
\begin{cases}
1, & \text{if }  [\rank(\x)]_i \le k \\
0, & \text{otherwise.}
\end{cases}.
\end{equation}
The \textbf{top-k} operator returns the values themselves if they are within the top-$k$ values or $0$ otherwise, i.e.,
\begin{equation}
\topk(\x) \coloneqq \x \circ \topkmask(\x),
\end{equation}
where $\circ$ denotes element-wise multiplication.
The \textbf{top-k in magnitude} operator is defined similarly as
\begin{equation}
\topkmag(\x) \coloneqq \x \circ \topkmask(|\x|).
\end{equation}

To illustrate, if $x_3 \ge x_1 \ge x_2$
and $|x_2| \ge |x_3| \ge |x_1|$, then
\begin{itemize}[topsep=0pt,itemsep=2pt,parsep=2pt,leftmargin=10pt]

\item $\argsort(\x) = (3, 1, 2)$

\item $\sort(\x) = (x_3, x_1, x_2)$

\item $\rank(\x) = (2, 3, 1)$

\item $\topkmask(\x) = (1, 0, 1)$

\item $\topk(\x) = (x_1, 0, x_3)$,

\item $\topkmag(\x) = (0, x_2, x_3)$,

\end{itemize}
where in the last three, we used $k=2$.

\paragraph{Permutahedron.} 

The permutahedron associated with a vector $\w \in \RR^n$, a well-known object in combinatorics \citep{bowman1972permutation,ziegler2012lectures},
 is the convex hull of the permutations of $\w$, i.e.,
\begin{equation}
P(\w) \coloneqq \conv(\{\w_\sigma \colon \sigma \in \Sigma\}) \subset \RR^n.
\end{equation}
We define the linear maximization oracles (LMO) associated with $P(\w)$ by
\begin{equation}
\begin{aligned}
f(\x, \w) &\coloneqq \max_{\y \in P(\w)} \langle \x, \y \rangle \\
\y(\x, \w) &\coloneqq \argmax_{\y \in P(\w)} ~ \langle \x, \y \rangle = \nabla_1 f(\x, \w),
\end{aligned}
\label{eq:lmo}
\end{equation}
where $\nabla_1 f(\x, \w)$ is technically a subgradient of $f$ w.r.t. $\x$.
The LMO can be computed in $O(n \log n)$ time. Indeed, the calculation of the LMO reduces to a sorting operation, as shown in the following known proposition. A proof is included for completeness in Appendix \ref{proof:lmo}.
\vspace{1em}
\begin{proposition}{(Linear maximization oracles)}\label{prop:lmo}

If $w_1 \ge \dots \ge w_n$ (if not, sort $\w$), then
\begin{equation}
f(\x, \w) = \sum_{i=1}^n w_i x_{[i]} 
\quad \text{and} \quad
\y(\x, \w) = \w_{\rank(\x)}.
\end{equation}
\end{proposition}

\paragraph{LP formulations.} 

Let us denote the reversing permutation by 
$\rhov \coloneqq (n, n-1, \dots, 1)$.
\citet{blondel2020fast} showed that the sort and rank operators can be formulated
as linear programs (LP) over the permutahedron:
\begin{equation}
\begin{aligned}
\sort(\x) &= \y(\rhov, \x) = \argmax_{\y \in P(\x)} \langle \rhov, \y \rangle \\
\rank(\x) &= \y(-\x, \rhov) = \argmax_{\y \in P(\rhov)} \langle -\x, \y \rangle.
\end{aligned}
\end{equation}
In the latter expression, the minus sign is due to the fact that we use the convention
that smaller rank indicates higher value (i.e., the maximum value has rank $1$).

\begin{figure}[t]
\centering
\includegraphics[scale=3.3]{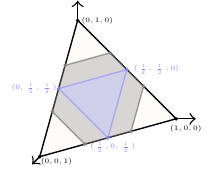}
\caption{The permutahedron $P(\w)$ is a polytope whose vertices are permutations of $\w$.
Depending, on the choice of $\w$, it can express several known polytopes.
When $\w = (1, 0, 0)$, $P(\w)$ is the probability simplex (light beige), which corresponds to the top-$1$ setting.
When $\w = (\frac{1}{2}, \frac{1}{2}, 0)$, $P(\w)$ is the capped probability simplex (blue), which corresponds to the top-$k$ setting (here, with $k=2$).
When $\w = (\frac{2}{3}, \frac{1}{3}, 0)$, $P(\w)$ is an hexagon,
which corresponds to the partial ranking setting (gray).
}
\label{fig:permutahedron}
\end{figure}

Although not mentioned by \citet{blondel2020fast}, it is also easy to express the top-$k$ mask operator as an LP
\begin{equation}
\topkmask(\x) = \y(\x, \ones_k) = \argmax_{\y \in P(\ones_k)} \langle \x, \y \rangle,
\label{eq:topk_mask}
\end{equation}
where $\ones_k \coloneqq (\underbrace{1, ..., 1}_{k}, \underbrace{0, \dots, 0}_{n-k})$. 
For this choice of $\w$, the permutahedron enjoys a particularly simple expression
\begin{equation}
P(\ones_k) = \{\y \in \RR^n \colon \langle \y, \ones \rangle = k, \y \in [0, 1]^n\}
\end{equation}
and $P(\ones_k / k)$ is known as the capped simplex \citep{warmuth2008randomized,blondel2020learning}.
This is illustrated in Figure \ref{fig:permutahedron}.
To obtain relaxed operators, \citet{blondel2020fast} proposed to introduce regularization in
\eqref{eq:lmo} (see ``recovering the previous framework'' in the next section) 
and used a reduction to isotonic optimization.

\section{Proposed generalized framework}\label{sec:proposed_framework}

In this section, we generalize the framework of \citet{blondel2020fast} by adding an optional nonlinearity $\varphi(\x)$.
In addition to the operators covered by the previous framework, this allows us to directly express the top-$k$ in magnitude operator, which was not possible before. We also support $p$-norm regularization, which allows to express differentiable \emph{and} sparse operators when $1 < p < 2$.

\paragraph{Introducing a mapping $\varphi$.}

Consider a  mapping 
$\varphi(\x) \coloneqq (\phi(x_1), \dots, \phi(x_n))$.
Given $\x \in \RR^n$ and $\w \in \RR^n$, we define
\begin{equation}
\begin{aligned}
f_\varphi(\x, \w) &\coloneqq f(\varphi(\x), \w) \\
\y_\varphi(\x, \w) &\coloneqq \nabla_1 f_\varphi(\x, \w).
\end{aligned}
\end{equation}
When $\varphi(\x) = \x$ (identity mapping), we clearly recover the existing framework, i.e., 
$f_\varphi(\x, \w) = f(\x, \w)$ and $\y_\varphi(\x, \w) = \y(\x, \w)$.

When $\varphi(\x) \neq \x$, our framework starts to differ from the previous one,
since $\varphi$ affects differentiation.
By the chain rule and Danskin's theorem (\citeyear{danskin1966theory}), 
we get
\begin{equation}
\begin{aligned}
\y_\varphi(\x, \w)
&\coloneqq \nabla_1 f_\varphi(\x, \w) \\
&= \partial \varphi(\x)^\top \y(\varphi(\x), \w) \\
&= (\phi'(x_1), \dots, \phi'(x_n)) \circ \y(\varphi(\x), \w),
\end{aligned}
\label{eq:chain_rule}
\end{equation}
where $\partial \varphi(\x) \in \RR^{n \times n}$ denotes the Jacobian of $\varphi(\x)$ and $\y(\x, \w)$ is given by Proposition \ref{prop:lmo}. 

\paragraph{Top-k in magnitude.}

As we emphasized, one advantage of our proposed generalization is that we can express
the top-$k$ in magnitude operator.
Indeed, with $\phi(x) = \frac{1}{2} x^2$, we can see from \eqref{eq:topk_mask} and \eqref{eq:chain_rule} that we have for all $\x \in \RR^n$
\begin{equation}
\topkmag(\x) = \y_\varphi(\x, \ones_k) = \nabla_1 f_\varphi(\x, \ones_k).
\label{eq:topkmag}
\end{equation}
Obviously, for all $\x \in \RR^n_+$, we also have $\topkmag(\x) = \topk(\x)$.
Top-k in magnitude is useful for pruning weights with small magnitude in a neural network,
as we demonstrate in our experiments in \S\ref{sec:experiments}.

\paragraph{Introducing regularization.}

We now explain how to make our generalized operator differentiable.
We introduce convex regularization $R : \RR^n \to \RR$ in the dual space:
\begin{equation}
f^*_{\varphi,R}(\y, \w) \coloneqq f_\varphi^*(\y, \w) + R(\y),
\end{equation}
where $f^*_\varphi$ is the conjugate of $f_\varphi$ in the first argument.
Going back to the primal space, we obtain a new relaxed operator. A proof is given in Appendix \ref{proof:relaxed_operator}.
\vspace{1em}
\begin{proposition}{(Relaxed operator)}\label{prop:relaxed_operator}

Let $R \colon \RR^n \to \RR$ be a convex regularizer.
Then
\begin{equation}
\begin{aligned}
f_{\varphi,R}(\x, \w) 
&\coloneqq \max_{\y \in \RR^n} \langle \y, \x \rangle - f_\varphi^*(\y, \w) - R(\y) \\
&= \min_{\u \in \RR^n} R^*(\x-\u) + f_\varphi(\u, \w) \\
\y_{\varphi,R}(\x, \w) 
&\coloneqq \y^\star = \nabla R^*(\x - \u^\star) = \nabla_1 f_{\varphi,R}(\x, \w).
\end{aligned}
\end{equation}
\end{proposition}
The mapping $\varphi$ also affects conjugacy. 
We have the following proposition.
\begin{proposition}{(Conjugate of $f_\varphi$ in the first argument)}

If $\phi$ is convex and $\w \in \RR_+^n$, then
\begin{equation}
f_\varphi^*(\y, \w) = \min_{\z \in P(\w)} D_{\phi^*}(\y, \z),
\end{equation}
where 
$D_f(\y, \z) \coloneqq\sum_{i=1}^n z_i f(y_i/z_i)$.
\label{prop:conjugate_unregularized}
\end{proposition}
A proof is given in Appendix \ref{proof:conjugate_unregularized}.
The function $(y_i, z_i) \mapsto y'_i \phi^*(y_i / z_i)$ is known
as the perspective of $\phi^*$ and is jointly convex when $z_i > 0$.
The function $D_f(\y, \z)$ is known as the $f$-divergence between $\y$ and $\z$. 
Therefore, $f_\varphi^*(\y, \w)$ can be seen as the minimum ``distance'' between $\y$ and
$P(\w)$ in the $\phi^*$-divergence sense.

\paragraph{Recovering the previous framework.}

If $\phi(x) = x$, then
\begin{equation}
\phi^*(y_i / z_i) =    
\begin{cases}
0, & \text{if }  y_i = z_i \\
\infty, & \text{otherwise}
\end{cases}.
\end{equation}
This implies that
$f_\varphi^*(\y, \w) = f^*(\y, \w) = \delta_{P(\w)}(\y)$, the indicator function of $P(\w)$,
which is $0$ if $\y \in P(\w)$ and $\infty$ otherwise.
In this case, we therefore obtain
\begin{equation}
f_{\varphi,R}(\x, \w) 
= \max_{\y \in P(\w)} \langle \y, \x \rangle - R(\y),
\end{equation}
which is exactly the relaxation of \citet{blondel2020fast}.

\paragraph{Differentiable and sparse top-k operators.} 

\begin{figure*}[ht]
 \centering
\includegraphics[width=1.\textwidth]{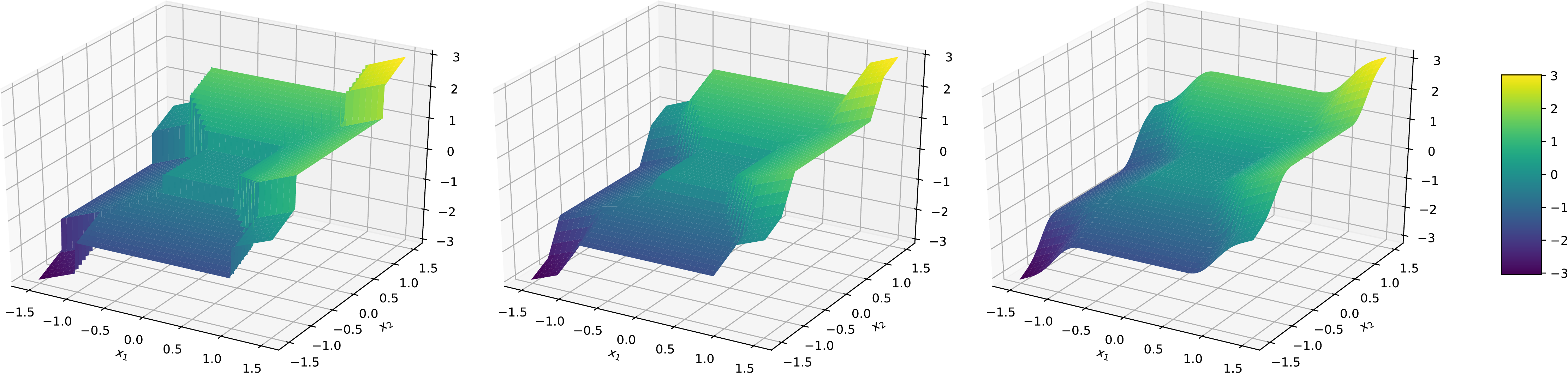}
\caption{\textbf{Example of our relaxed top-k operators.} We take $\phi(x) = \frac12 x^2$ and $k=2$. For an input $\x = (x_1, x_2, \frac{1}{2}, 1)$, we plot $y_{\varphi,R}(\x, 1_k)_1 + y_{\varphi,R}(\x, 1_k)_2$ for $R=0$ (left), $R=\frac\lambda 2\|\x\|^2_2$ (center) and $R=\frac\lambda p\|\x\|^p_p$ with $p=\frac43$ (right). We take $\lambda = 0.3$. While no regularization leads to a discontinuous mapping, the $2$-norm regularization leads to continuity and a.e. differentiability, and the ${\frac43}$-norm regularization provides a continuously differentiable mapping. We emphasize that, although in the left plot the graph looks connected, it is actually a discontinuous function. Note that our relaxed operators are \textbf{sparse} as they are exactly $0$ in the center. }
\label{fig:topk}
 \end{figure*}

To obtain a relaxed top-$k$ operator with our framework, we simply replace $f_\varphi$
with $f_{\varphi,R}$ and $\y_\varphi$ with $\y_{\varphi,R}$ in \eqref{eq:topkmag} to define
\begin{equation}
\topkmag_R(\x) \coloneqq \y_{\varphi,R}(\x, \ones_k) = \nabla_1 f_{\varphi,R}(\x, \ones_k).
\end{equation}
A relaxed top-$k$ mask can be defined in a similar way,
but using $\phi(x) = x$ instead of $\phi(x) = \frac{1}{2} x^2$.
For the regularization $R$, we propose to use $p$-norms to the power $p$:
\begin{equation}
R(\y) = \frac{1}{p} \|\y\|_p^p \coloneqq \frac{1}{p} \sum_{i=1}^n |y_i|^p.    
\end{equation}
The choice $p=2$ used in previous works leads to sparse outputs but is not differentiable everywhere.
Any $p$ between $1$ (excluded) and $2$ (excluded) leads to differentiable \emph{and} sparse outputs.
We propose to use $p=4/3$ to obtain a differentiable everywhere operator, and $p=2$ to obtain a differentiable a.e. operator, which is more convenient numerically.
This is illustrated in Figure~\ref{fig:topk}. 

\paragraph{Connection with $k$-support and OWL norms.} 

When $\phi(x) = \frac{1}{2} x^2$ and $\w = \ones_k$, we obtain
\begin{equation}
f_\varphi^*(\y, \w) = \frac{1}{2} \min_{\z \in [0, 1]^n} \sum_{i=1}^n \frac{y_i^2}{z_i}
\quad \text{s.t.} \quad
\langle \z, \ones \rangle = k,
\end{equation}
which is known as the squared k-support norm \cite{argyriou2012sparse, mcdonald2014spectral, eriksson2015k}. 
Our formulation is a generalization of the squared $k$-support norm, 
as it supports other choices of $\w$ and $\phi$. 
For instance, we use it to define a new notion of $k$-support negentropy in Appendix
\ref{app:k_support_negent}. When $\varphi(\x) = |\x|$, we recover the ordered weighted lasso (OWL) norm \citep{zeng2014ordered} as
\begin{equation}
f_\varphi(\x, \w) = \sum_{i=1}^n w_i |x|_{[i]}.
\end{equation}
With that choice of $\varphi$, it is easy to see from \eqref{eq:chain_rule} that \eqref{eq:topkmag} becomes
a \emph{signed} top-$k$ mask. Note that, interestingly, $k$-support and OWL norms are not defined in the same space.

\paragraph{Biconjugate interpretation.} 

Let us define the set of $k$-sparse vectors, which is nonconvex, as 
$S_k \coloneqq \{\x \in \RR^n \colon \|\x\|_0 \leq k\},$ where $\|\x\|_0$ is the number of non zero elements in $\x$.
We saw that if $\phi(x) = \frac12 x^2$ then $f_\varphi^*(\y, \ones_k)$ is the squared $k$-support norm. It is known to be 
the biconjugate (i.e., the tightest convex relaxation) of the squared $L_2$ norm restricted to $S_k$ 
\citep{eriksson2015k,sparse_ot}. We now prove a more general result: $f_\varphi^*(\y, \ones_k)$ is the biconjugate of $\sum_{i=1}^n\phi^*(y_i)$ restricted to $S_k$.
\begin{proposition}{(Biconjugate interpretation)}\label{prop:biconjugate}

Let $\Phi_k(\y) \coloneqq \sum_{i=1}^n\phi^*(y_i) + \delta_{S_k}(\y)$. Suppose that $\phi$ is convex.
Then the biconjugate of $\Phi_k$ is given by
\begin{equation}
\Phi^{**}_k(\y) = f_\varphi^*(\y, \ones_k)
\end{equation}
\end{proposition}
See Appendix \ref{proof:biconjugate} for a proof.

\section{Algorithms}
\label{sec:algorithms}

In this section, we propose efficient algorithms for computing our operators. We first show that the calculation of our relaxed operator reduces to isotonic optimization.

\paragraph{Reduction to isotonic optimization.}

We now show how to compute $\u^\star$ in Proposition \ref{prop:relaxed_operator}
by reduction to isotonic optimization,
from which $\y^\star$ can then be recovered by $\y^\star = \nabla R^*(\x - \u^\star)$.
We first recall the case $\varphi(\x) = \x$, which was already proved in existing works 
\citep{lim2016efficient,blondel2020fast}. 
\begin{proposition}{(Reduction, $\varphi(\x) = \x$ case)}
\label{prop:iso_1}

Suppose that $R(\y) = \sum_{i=1}^n r(y_i)$.
Let $\sigma$ be the permutation sorting $\x$,
$\s \coloneqq \x_\sigma$
and
$$ 
\v^\star = \argmin_{v_1 \ge \dots \ge v_n} R^*(\s - \v) + f(\v, \w).
$$
Then $\u^\star$ from Proposition \ref{prop:relaxed_operator} is given by
$\u^\star = \v^\star_{\sigma^{-1}}$.  
\end{proposition}   
The set $\{\v \in \RR^n \colon v_1 \ge \dots \ge v_n\}$ is called the monotone cone.
Next, we show that a similar result is possible when $\phi(\x)$ and $r^*$ are both even functions (sign-invariant) and 
increasing on $\RR_+$. 
\begin{proposition}{(Reduction, $\varphi(\x)=\varphi(-\x)$ case})
\label{prop:iso_2}

Suppose that $R(\y) = \sum_{i=1}^n r(y_i)$ and $\varphi(\x) = (\phi(x_1), \dots, \phi(x_n))$.
Assume $\phi$ and $r^*$ are both even functions (sign-invariant) and 
increasing on $\RR_+$.
Let $\sigma$ be the permutation sorting $|\x|$,
$\s \coloneqq |\x|_\sigma$
and
$$ 
\v^\star = \argmin_{v_1 \ge \dots \ge v_n \ge 0} R^*(\s - \v) + f_\varphi(\v, \w).
$$
Then, $\u^\star$ (Proposition \ref{prop:relaxed_operator}) is equal to $\mathrm{sign}(\x) \circ \v^\star_{\sigma^{-1}}$. 
\end{proposition}
See Appendix \ref{proof:example_isotonic} for a proof. 
Less general results are proved in \citep{zeng2014ordered,eriksson2015k} for specific cases of $\phi$ and $R$.
The set $\{v \in \RR^n \colon v_1 \ge \dots \ge v_n \ge 0\}$ is called the non-negative monotone cone.
In practice, the additional non-negativity constraint is easy to handle: we can solve the isotonic optimization problem without it
and truncate the solution if it is not non-negative \citep{nemeth2012project}.

\paragraph{Pool adjacent violator (PAV) algorithms.}

Under the conditions of Proposition \ref{prop:iso_2},
assuming $\v$ and $\w$ are both sorted, we have
from Proposition \ref{prop:lmo} that 
\begin{equation}
f(\v, \w) = \sum_{i=1}^n w_i v_i,
\quad
f_\varphi(\v, \w) = \sum_{i=1}^n w_i \phi(v_i).
\end{equation}
We then get that the problems in Proposition \ref{prop:iso_1} and \ref{prop:iso_2} are coordinate-wise separable:
\begin{equation}
\v^\star = \argmin_{v_1 \ge \dots \ge v_n} \sum_{i=1}^n h_i(v_i),
\label{eq:isotonic_separable}
\end{equation}
for $h_i(v_i) = r^*(s_i - v_i) + w_i \phi(v_i)$.
Such problems can be solved in $O(n)$ time using
the pool adjacent violator (PAV) algorithm \citep{best2000minimizing}.
This algorithm works by partitioning the set $[n]$ into disjoint sets $(B_1, \cdots B_m)$, starting from $m=n$ and $B_i = \{ i\}$, and by merging these sets until the isotonic condition is met. A pseudo-code is available for completness in Appendix \ref{app:pav}. At its core, PAV simply needs a routine to solve the ``pooling'' subproblem
\begin{equation}
\gamma^\star_B = \argmin_{\gamma \in \RR}\sum_{i \in B} h_i(\gamma)
\label{eq:pav_subproblem}
\end{equation}
for any $B \subseteq [n]$. Once the optimal partition $(B_1, \cdots B_m)$ is identified,
we have that 
\begin{equation}
\v^\star =  (\underbrace{\gamma_{B_1}^\star, \dots, \gamma_{B_1}^\star}_{|B_1|}, \cdots, \underbrace{\gamma_{B_m}^\star, \dots, \gamma_{B_m}^\star}_{|B_m|}) \in \RR^n.
\end{equation}
Because Proposition \ref{prop:iso_1} and \ref{prop:iso_2} require to obtain
the sorting permutation $\sigma$ beforehand, the total time complexity for our operators
is $O(n \log n)$.

\paragraph{Example.} 

Suppose $\phi(x) = \frac{1}{2} x^2$ and $R = \frac{\lambda}{2} \|.\|^2_2$, where $\lambda > 0$. 
Since  $R^* = \frac{1}{2 \lambda} \| .\|^2$,
this gives
$h_i(v_i) = \frac12 (w_i v^2_i + \frac1\lambda(s_i - v_i)^2)$. 
The solution of the sub-problem is then
$$
\gamma^\star_B = \frac{\sum_{i\in B}s_i}{\sum_{i\in B}(\lambda w_i + 1)}.
$$
Using this formula, when $\lambda$ is small enough, we can upper-bound the error between the hard and relaxed operators: ${\|\topkmag_{R}(\x) - \topkmag(\x)\|_{\infty}} \leq \lambda \|\x\|_{\infty}$. See Appendix \ref{proof:example_isotonic} for details and for the case $p = \frac43$.

\paragraph{Dykstra's alternating projection algorithm.} 

The PAV algorithm returns an exact solution of \eqref{eq:isotonic_separable} in $O(n)$ time.
Unfortunately, it relies on element-wise dynamical array assignments, which makes it potentially slow on GPUs and TPUs. We propose an alternative to obtain faster computations.
Our key insight is that by defining
\begin{equation}
\begin{aligned}
C_1 &\coloneqq \{\v \in \RR^n \colon v_1 \ge v_2, v_3 \ge v_4, \dots \} \\
C_2 &\coloneqq \{\v \in \RR^n \colon v_2 \ge v_3, v_4 \ge v_5, \dots \},
\end{aligned}
\end{equation}
one has
$
\{\v \in \RR^n \colon v_1 \ge \dots \ge v_n\} = C_1 \cap C_2.
$
We can therefore rewrite \eqref{eq:isotonic_separable} as
\begin{equation}
\v^\star = \argmin_{\v \in C_1 \cap C_2} \sum_{i=1}^n h_i(v_i).
\end{equation}
In the case $R = \frac{1}{2} \|.\|^2$ and $\phi(x) = x$ or $\frac12 x^2$,
this reduces to a projection onto $C_1 \cap C_2$,
for which we can use Dykstra's celebrated projection algorithm \cite{boyle1986method, combettes2011proximal}. \citet{jegelka2013reflection} have used this method for computing the projection onto the intersection of submodular polytopes, whereas we project onto a single polyhedral face. When $\phi(x) = x$, Dykstra's projection algorithm takes the simple form in Algorithm \ref{alg:dykstra}.

\begin{figure}[H]
    \begin{minipage}[c]{0.65\linewidth}
        \includegraphics[width=\textwidth]{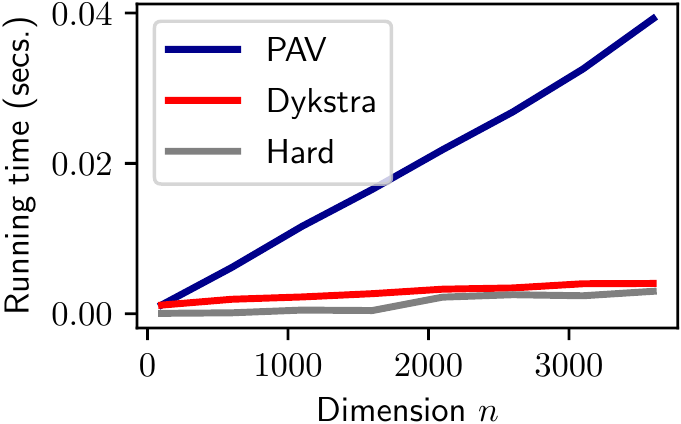}
    \end{minipage}\hfill
    \begin{minipage}[c]{0.35\linewidth}
        \vspace{-1em}
\caption{\textbf{Runtime comparison} for computing our relaxed top-$k$ on a TPU using PAV, Dykstra, as a function of the dimension $n$. For each $n$, we set $k = \lceil n/10 \rceil$. We also compare with the hard top-k computation.}\label{fig:timing_dykstra}
    \end{minipage}
    \vspace{-1em}
\end{figure}

\begin{algo}{(Dykstra's projection algorithm)
}\label{alg:dykstra}

Starting from $\v^0 = \s$, $\p^0 = \q^0 = \zeros$, Dykstra's algorithm iterates
$$
\begin{aligned}
&\y^k = \argmin_{\y \in C_1} \frac12 \|\v^k + \p^k - \y\|^2_2 + \langle \y, \w \rangle \\
&\p^{k+1} = \v^k + \p^k - \y^k  \\
&\v^{k+1} = \argmin_{\v \in C_2} \frac12 \|\y^k + \q^k - \v\|^2_2 + \langle \v, \w \rangle \\
&\q^{k+1} = \y^k + \q^k - \v^{k+1}.
\end{aligned}
$$
\end{algo}
Each argmin calculation corresponds to a Euclidean projection onto $C_1$ or $C_2$, which can be computed in closed form. Therefore, $\v^k$ provably converges to $\v^\star$. Each iteration of Dykstra's algorithm is learning-rate free, has linear time complexity and can be efficiently written as a matrix-vector product using a mask, which makes it particularly appealing for GPUs and TPUs. Interestingly, we find out that when $\w = \ones_k$, then Dykstra's algorithm converges surprisingly fast to the exact solution. We validate that Dykstra leads to a faster runtime than PAV in Figure \ref{fig:timing_dykstra}. We use $100$ iterations of Dykstra, and verify that we obtain the same output as PAV. 

For the general non-Euclidean $R^*$ case, i.e., $p \neq 2$,
we can use block coordinate ascent in the dual of \eqref{eq:isotonic_separable}. 
We can divide the dual variables into two blocks, corresponding to $C_1$ and $C_2$;
see Appendix \ref{proof:isotonic_reg_dual}.
In fact, it is known that in the Euclidean case, Dykstra's algorithm in the primal
and block coordinate ascent in the dual are equivalent \citep{tibshirani_2017}.
Therefore, although a vectorized implementation could be challenging for $p \neq 2$,
block coordinate ascent can be seen as an elegant way to generalize Dykstra's algorithm.
Convergence is guaranteed as long as each $h_i$ is strictly convex.

\paragraph{Differentiation.}

The Jacobian of the solution of the isotonic optimization problems can be expressed in closed form for any $p$-norm regularization.
\begin{proposition} \label{prop:jacobian}
(Differentiation)

Let $\v^\star =  (\gamma_{B_1}^\star, \dots, \gamma_{B_1}^\star \cdots, \gamma_{B_m}^\star, \dots, \gamma_{B_m}^\star)$ be the optimal solution of the isotonic optimization problem with $R = \frac1p \|.\|^p_p$ and $p > 0$.
Then one has that $\v^\star$ is differentiable with respect to $\s = (s_1, \dots, s_n)$. Furthermore, for any $r \in \{1, \dots, m\}$ and $i \in B_r$,
\begin{equation}
\frac{\partial \gamma^\star_{B_r}}{\partial s_i} =    
\begin{cases}
\frac{|\gamma^\star_{B_r} - s_i|^{q-2}}{\sum_{j\in {B_r}} |\gamma^\star_{B_r} - s_j|^{q-2}} & \text{if } \phi(x) = x \\
& \\
 \frac{(q-1)|\gamma^\star_{B_r} - s_i|^{q-2}}{\sum_{j\in {B_r}} (q-1)|\gamma^\star_{B_r} - s_j|^{q-2} + w_j}& \text{if } \phi(x) = \frac12 x^2
\end{cases}
\end{equation}
where $q$ is such that $\frac1p + \frac1q = 1$. When $i \notin B_r$, one simply has $\frac{\partial \gamma^\star_{B_r}}{\partial s_i} = 0.$ 
\end{proposition}
One then has $\partial v_j^\star /  \partial s_i =  \partial \gamma^\star_{B_{r_j}}/{\partial s_i}$ where $r_j$ is such that $v^\star_j = \gamma^\star_{B_{r_j}}$. See Appendix \ref{proof:jacobian} for a proof. Thanks to Proposition \ref{prop:jacobian}, we do not need to solve a linear system to compute the Jacobian of the solution $\v^\star$, in contrast to implicit differentiation of general optimization problems \cite{blondel2021efficient}. In practice, this also means that we do not need to perform backpropagation through the unrolled iteratations of PAV or Dykstra's projection algorithm to obtain the gradient of a scalar loss function, in which our operator is incorporated. In particular, we do not need to store the intermediate iterates of these algorithms in memory. Along with PAV and Dykstra's projection algorithm, we implement the corresponding Jacobian vector product routines in JAX, using Proposition \ref{prop:jacobian}.

\section{Experiments}\label{sec:experiments}

We now demonstrate the applicability of our top-k operators through experiments. 
Our JAX \cite{jax2018github} implementation is available at the following \href{https://github.com/google-research/google-research/tree/master/sparse_soft_topk}{URL}. See Appendix \ref{app:exp_details} for additional experimental details.

\paragraph{Weight pruning in neural networks.} 

\begin{figure}[h]
\centering
\includegraphics[width=0.99\columnwidth]{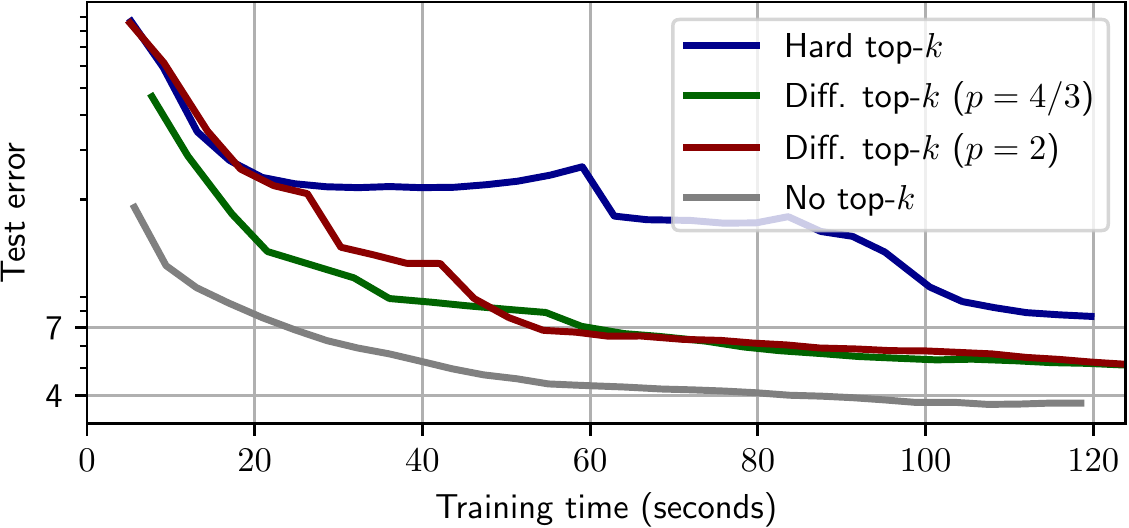} 
\caption{\textbf{Test error} with respect to training time when training an MLP on MNIST. We compare the baseline (grey) with the case where $90 \%$ of the weights are set to 0 by magnitude pruning, using a differentiable a.e. (red), fully differentiable (green) or a hard top-$k$ (blue).}\label{fig:mnist}
\vspace{-1em}
\end{figure}
We experimentally validate the advantage of using a smoothed top-k for
weight pruning in neural networks. We use a multilayer perceptron (MLP) with 2 hidden layers
and with ReLU activation. The width of the layers are respectively $784$, $32$, $32$ followed by a linear classification head of width $10$. More precisely, our model takes as input an image $\a \in \RR^{784}$ and outputs
$$
\x = W_3 \sigma(W_2\sigma(W_1\a + \b_1) + \b_2) + \b_3
\quad \text{(logits)},
$$
where $W_1 \in \RR^{32 \times 784}$, $\b_1 \in \RR^{32}$, $W_2 \in \RR^{32 \times 32}$, $\b_2 \in \RR^{32}$, $W_1 \in \RR^{10 \times 32}$, $\b_3 \in \RR^{10}$ and $\sigma$ is a ReLU. In order to perform
weight pruning we parametrize each $W_i$ as 
$
W_i = \mathrm{topkmag}_R(W'_i)
$
and learn $W'_i$ instead of learning $W_i$ directly.
The output is then fed into a cross-entropy loss.
We compare the performance of the model when applying a hard vs differentiable top-k operator 
to keep only $10 \%$ of the coefficients. For the differentiable top-k, we use a regularization $R(\y) = \frac{\lambda}{p} \|\y\|^p$ with $p \in \{\frac43, 2\}$ and $\lambda = 10^{-4}.$ 
We find out that the model trained with the differentiable top-k trains significantly faster than the one trained with the hard top-k. We also verify that our relaxed top-k maintains the $10 \%$ rate of non-zero weights. Results on MNIST are displayed in Figure \ref{fig:mnist}. We also compare with an entropy-regularized approximation of the top-k operator using the framework proposed in \citet{cuturi2019differentiable} and adapted in \citet{petersen2022differentiable}. To guarantee the sparsity of the weights, we use the "straight-through" trick: the hard top-k is run on the forward pass but we use the gradient of the relaxed top-k in the backward pass. This method leads to a test error of $5.9 \%$, which is comparable to the results obtained with our differentiable operators.

\paragraph{Smooth top-k loss.} 

\begin{figure}[H]
\centering
\includegraphics[width=1\columnwidth]{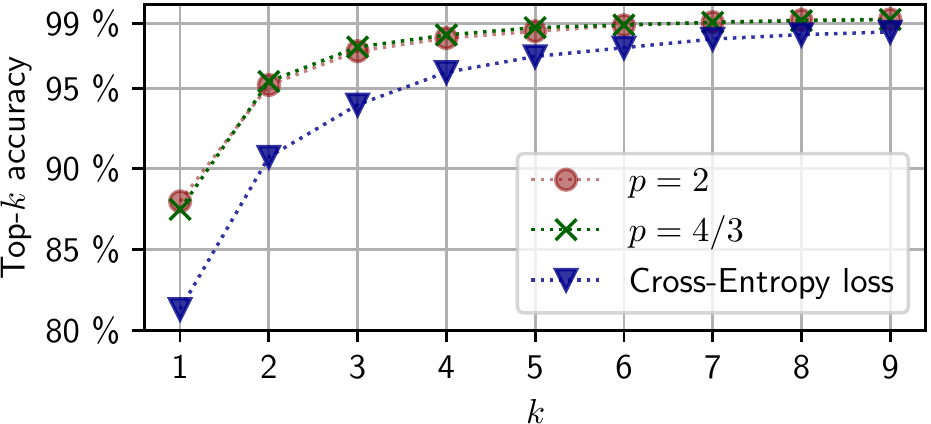} 
\caption{\textbf{Validation top-k accuracy} when fine-tuning a ViT-B/16 on CIFAR 100 using either the cross-entropy loss or our smooth top-$3$ loss for training.  We have the following running times obtained with a TPUv3-8.
Baseline: 9.5 sec/step,
$p=2$: 9.7 sec/step and
$p=4/3$: 10 sec/step.}\label{fig:cifar}
\vspace{-1em}
\end{figure}

To train a neural network on a classification task, one typically minimizes the cross-entropy loss, whereas the performance of the network is evaluated using a top-$k$ test accuracy. There is therefore a mismatch between the loss used at train time and the metric used at evaluation time. \citet{cuturi2019differentiable} proposed to replace the cross-entropy loss with a differentiable top-$k$ loss. 
In the same spirit, we propose to finetune a ViT-B/16 \cite{dosovitskiy2020image} pretrained on the ImageNet21k dataset on CIFAR 100 \cite{krizhevsky2009learning} using a smooth and sparse top-$k$ loss instead of the cross-entropy loss.
We use a Fenchel-Young loss \cite{blondel2020learning}. It takes as input the vector $\a$ and parameters $\theta$ of a neural network $g_\theta$:
$$
\begin{aligned}
\x &= g_\theta(\a) \quad \text{(logits)} \\
\ell(\x, \t) &=  f_{\varphi,R}(\x, \ones_k) - 
\langle \x, \t \rangle,
\end{aligned}
$$
where $f_{\varphi,R}(\x, \ones_k)$ is given by Proposition \ref{prop:relaxed_operator} and $\t$ is a one-hot encoding of the class of $\a$. We set $\phi(x) = x$ as we want a top-$k$ mask. We consider $p$ norm regularizations for $R$, where $p=2$ or $p = 4/3$. We take $k=3$.
We use the exact same training procedure as described in \citet{dosovitskiy2020image} and use the corresponding pretrained ViT model B/16, and train our model for $100$ steps. Results are reported in Figure \ref{fig:cifar}. We find that the ViT finetuned with the smooth top-$3$ loss outperforms the one finetuned with the cross-entropy loss in terms of top-$k$ error, for various $k$. 

\paragraph{Sparse MoEs.}

\begin{figure}[H]
    \begin{minipage}[c]{0.6\linewidth}
        \includegraphics[width=\textwidth]{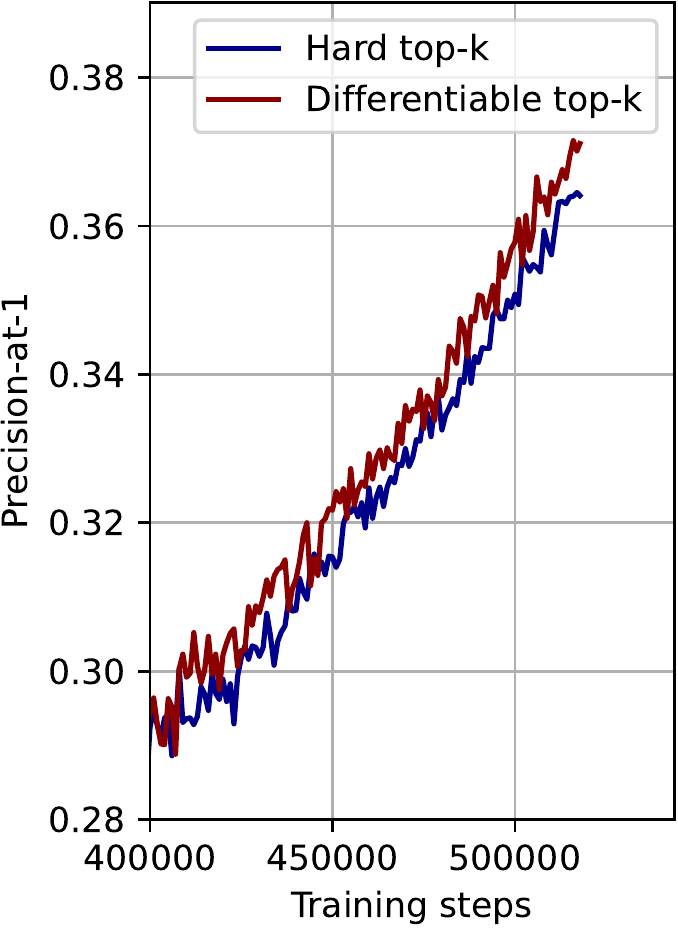}
    \end{minipage}\hfill
    \begin{minipage}[c]{0.33\linewidth}
        \vspace{-1em}
\caption{\textbf{Precision-at-1} on the JFT-300M dataset when using a hard top-$k$ (baseline, in blue) or a differentiable a.e. one (in red) in a sparse MoE with a ViT-S/32. We zoom in on the last training steps, where our proposed method outperforms the baseline. The runtime is 10 hours for the baseline and 15 for the differentiable a.e. top-k (gradient calculation is the bottleneck here).
}\label{fig:moe}
    \end{minipage}
    \vspace{-1em}
\end{figure}

Finally, we demonstrate the applicability of our proposed smooth top-$k$ operators on a large-scale classification task using vision sparse mixture of experts (V-MoE) \cite{riquelme2021scaling}. Vision transformers (ViTs) are made of a succession of self-attention layers and MLP layers. The idea of V-MoEs is to replace MLPs in ViTs by a sparsely-gated mixture of MLPs called experts. This way, only some of the experts are activated by a given patch token. At the heart of the token-expert assignment lies a routing mechanism which performs a top-$k$ operation on gate values. We focus on the MoE with expert choice routing framework \cite{zhou2022mixture}, where each expert is assigned to $k$ tokens. We train a S/32 variant of the V-MoE model, with $32 \times 32$ patches on the JFT-300M dataset
\cite{sun2017revisiting}, a dataset with more than 305 million images. Our model has $32$ experts, each assigned to $k=28$ tokens selected among $n=400$ at each MoE layer. We compare the validation accuracy when using the baseline (hard top-$k$) with our relaxed operator. We use $p=2$ and Dykstra's projection algorithm, as we found it was the fastest method on TPU. 
We used the training procedure proposed by \citet{zhou2022mixture} to obtain a fair comparison with the baseline. Due to the large size of the JFT-300M dataset (305 million images), we performed one run, as in \citet{sparse_ot}. 
We find that our approach improves validation performance. Results are displayed in Figure \ref{fig:moe}.

\section{Discussion}

\paragraph{Advantage of the non-linearity.}

As an alternative to performing a relaxed top-k operator in magnitude of $\x$, one can perform a differentiable top-k mask on $|\x|$, and then multiply the output by $\x$. This alternative would also lead to a differentiable top-k operator in magnitude. However, our operator has more principled behavior at the limit cases. For instance, as $\lambda \to \infty$, it is easy to see that the relaxed top-k mask converges to the vector $(k / n) \times 1_n$. Therefore, a rescaling by $n/k$ is needed to obtain the identity as $\lambda \to \infty$, in contrast to our top-k in magnitude. From a theoretical point of view, the introduction of a non-linearity allows us to draw connections with the $k$-support norm. It also has a bi-conjugate interpretation, which we believe has an interest by itself.

\paragraph{Sensitivity to the choice of p.}

\begin{figure}[h]
\centering
\includegraphics[width=0.99\columnwidth]{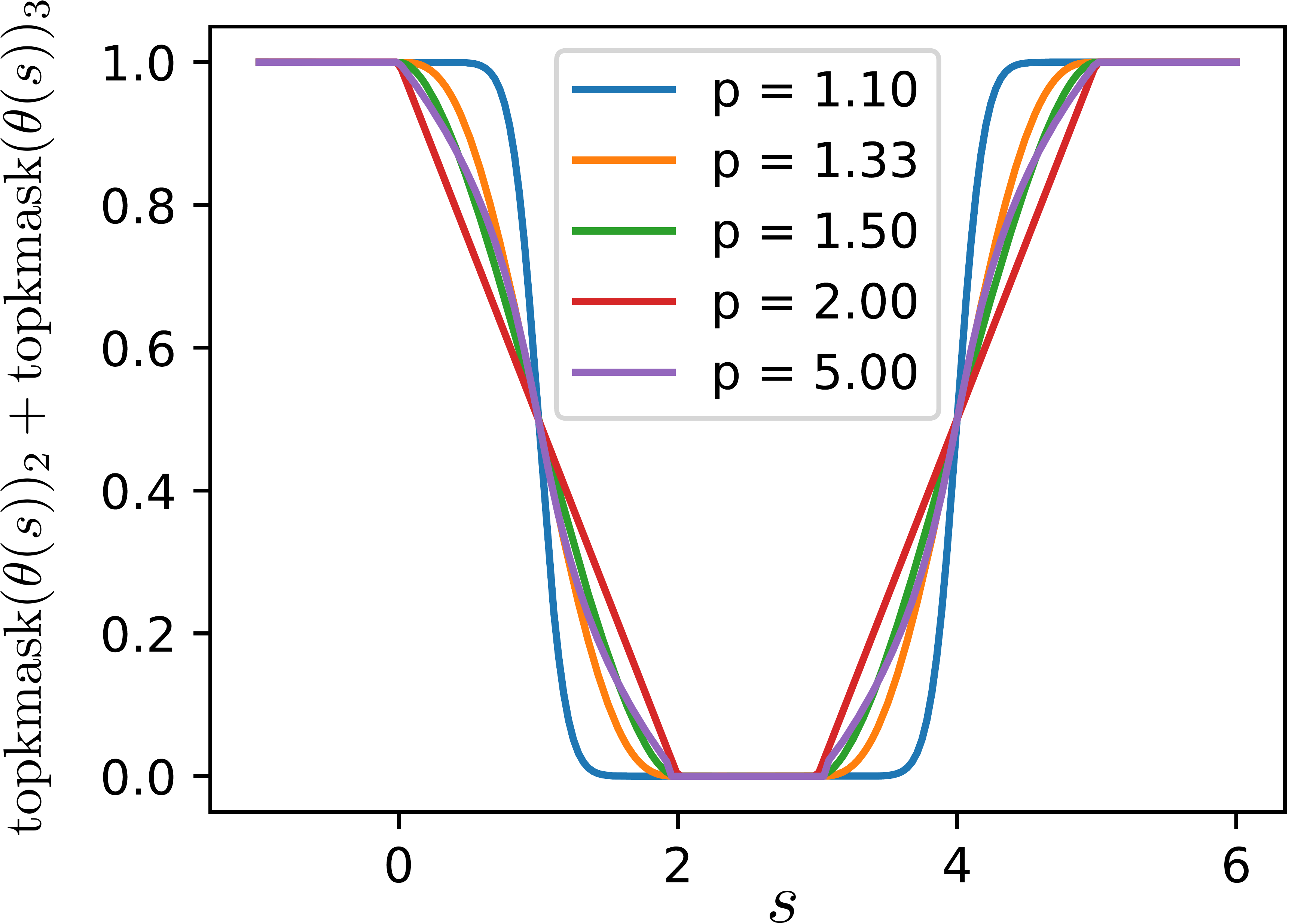} 
\caption{\textbf{Illustration of our differentiable and sparse top-k mask.} Same setup as for \autoref{fig:intro}, with more values for $p$.}\label{fig:multiple_p}
\end{figure}

The subproblem needed within PAV enjoys a closed form only for specific choices of $p$. This is why we focused on $p=2$ and $p=\frac43$ in our experiments. However, we stress out that the proposed methods work for any choice of $p$. As an example, we provide the same illustration as for \autoref{fig:intro} in \autoref{fig:multiple_p}.
\section{Conclusion}

In this work, we proposed a generalized framework to obtain fast, differentiable (or differentiable a.e.) and sparse top-$k$ and top-$k$ masks operators, including operators that select values in magnitude. Thanks to a reduction to isotonic optimization, we showed that these operators can be computed using either the Pool Adjacent Violators (PAV) algorithm or Dykstra's projection algorithm, the latter being faster on TPU hardware. We successfully demonstrated the usefulness of our operators for weight pruning, top-$k$ losses and as routers in vision sparse mixture of experts.

\paragraph{Acknowledgments.}

We thank Vincent Roulet and Joelle Barral for comments on a draft of this paper. We
thank Felipe Llinares-López for helpful feedbacks regarding the experiments, as well as Fabian Pedregosa for fruitful mathematical discussions. We also thank the anonymous reviewers for their feedback.

\bibliography{references}
\bibliographystyle{icml2023}

\newpage
\appendix
\onecolumn

\section{Proofs}

\subsection{Linear Maximization Oracle - Proof of Proposition \ref{prop:lmo}}
\label{proof:lmo}

Assuming $\w$ is sorted in descending order, we have for any $\x \in \RR^n$
\begin{equation}
\begin{aligned}
\sum_{i=1}^n w_i x_{[i]}
&= \max_{\sigma \in \Sigma} \langle \x_\sigma, \w \rangle \\
&= \max_{\pi \in \Sigma} \langle \x, \w_\pi \rangle \\
&= \max_{\y \in P(\w)} \langle \x, \y \rangle.
\end{aligned}
\end{equation}
In the first line, we used that the inner product is maximized by finding the permutation $\sigma$ sorting $\x$ in descending order. In the second line, we used that
$\langle \x_\sigma, \w \rangle = \langle \x, \w_\pi \rangle$, if $\pi$ is the inverse permutation of $\sigma$. In the third line, we used the fundamental theorem of linear programming,
which guarantees that the solution happens at one of the vertices of the polytope.
To summarize, if $\sigma$ is the permutation sorting $\x$ in descending order,
then $\y^\star = \w_{\sigma^{-1}}$.

\subsection{Relaxed operator - Proof of Proposition \ref{prop:relaxed_operator}}
\label{proof:relaxed_operator}
Recall that $f^*_{\varphi,R}(\y, \w) \coloneqq f^*_\varphi(\y, \w) + R(\y)$. 
We then have
\begin{equation}
f_{\varphi,R}(\x, \w) = \max_{\y \in \RR^n} \langle \y, \x \rangle - f_\varphi^*(\y, \w) - R(\y).
\end{equation}
It is well-known that if $h_1$ and $h_2$ are two convex functions,
then $(h_1 + h_2)^*$ is equal to the infimal convolution of $h_1^*$ with $h_2^*$
\citep[Theorem 4.17]{beck2017first}:
\begin{equation}
(h_1 + h_2)^*(\x) = (h_1^* \square h_2^*)(\x) \coloneqq \min_{\u \in \RR^n} h_1^*(\u) + h_2^*(\x - \u).
\end{equation}
With $h_1 = f_\varphi^*$ and $h_2 = R$, we therefore get
\begin{equation}
f_{\varphi,R}(\x, \w) = \min_{\u \in \RR^n} R^*(\x-\u) + f_\varphi(\u, \w).
\end{equation}
Finally, the expression of $\y^{\star}$
 follows from Danskin's theorem applied. 
\subsection{Conjugate - Proof of Proposition \ref{prop:conjugate_unregularized}}
\label{proof:conjugate_unregularized}

We have
\begin{equation}
\begin{aligned}
f^{*}_\varphi(\y,\w) 
&= \max_{\x \in \RR^n} \langle \x, \y \rangle - f_\varphi(\x, \w) \\
&= \max_{\x \in \RR^n} \langle \x, \y \rangle 
- \max_{\y' \in P(\w)} ~ \langle \varphi(\x), \y' \rangle \\
&= \max_{\x \in \RR^n} \min_{\y' \in P(\w)} \langle \x, \y \rangle - \langle \varphi(\x), \y' \rangle.
\end{aligned}    
\end{equation}
If $\w \in \RR_+^n$, then $\y' \in \RR^n_+$ for all $\y' \in P(\w)$.
Then the function 
$(\x, \y') \mapsto \langle \x, \y \rangle - \langle \varphi(\x), \y' \rangle$
is concave-convex and we can switch the min and the max to obtain
\begin{equation}
\begin{aligned}
f^{*}_\varphi(\y,\w) 
&= \min_{\y' \in P(\w)} \max_{\x \in \RR^n} \langle \x, \y \rangle - 
\langle \varphi(\x), \y' \rangle \\
&= \min_{\y' \in P(\w)} \sum_{i=1}^n y'_i \phi_i^*(y_i / y'_i).
\end{aligned}
\end{equation}

\subsection{Biconjugate interpretation - Proof of Proposition \ref{prop:biconjugate}}\label{proof:biconjugate}
One has 
$$
\Phi^{*}_k(\x) = \max_{\y \in S_k} \langle \x, \y \rangle - \sum^n_{i=1} \phi^*(y_i).
$$
As in \citep{kyrillidis2013sparse}, let $\Sigma_k$ be the set of subsets of $[n]$ with cardinality smaller than $k$.
Then 
$$
\Phi^{*}_k(\x) = \max_{I \subset \Sigma_k} \max_{\y \in \RR^n}  \sum_{i \in I} x_i y_i - \phi^*(y_i).
$$
This gives 
$$
\Phi^{*}_k(\x) = \max_{I \subset \Sigma_k} \sum_{i \in I} \phi(x_i) = \sum_{i=1}^k \varphi(\x)_{[i]}.
$$
Taking the conjugate gives the desired result.

\subsection{Reduction to isotonic optimization}\label{proof:reduction_iso}

We focus on the case when $\varphi$ is sign-invariant, i.e., 
$\varphi(\x) = \varphi(-\x)$, since the case $\varphi(\x) = \x$
is already tackled in \citep{lim2016efficient,blondel2020fast}.

We first show that $\u^\star$ preserves the sign of $\x$.
We do so by showing that for any $\u \in \RR^n$,
$\u' \coloneqq \mathrm{sign}(\x) \circ |\u|$ achieves smaller objective value than $\u$.
Recall that
\begin{equation}
f_{\varphi,R}(\x, \w) 
= \min_{\u \in \RR^n} R^*(\x-\u) + f_\varphi(\u, \w)
= \min_{\u \in \RR^n} R^*(\x-\u) + f(\varphi(\u), \w).
\end{equation}
Clearly, we have $f(\varphi(\u'), \w) = f(\varphi(\u), \w)$.
Moreover, 
if $R(\y) = \sum_{i=1}^n r(y_i)$,
then $R^*(\x - \u) = \sum_{i=1}^n r^*(x_i - u_i)$.
If $r^*$ is sign-invariant and increasing on $\RR_+$, we then have
\begin{equation}
\begin{aligned}
r^*(x_i - u'_i)
&= r^*(\mathrm{sign}(x_i)(|x_i| - |u_i|)) \\
&= r^*(|x_i| - |u_i|) \\
&\le r^*(x_i - u_i),
\end{aligned}
\end{equation}
where we used the reverse triangle inequality
$||x_i - |u_i|| \le |x_i - u_i|$.
We conclude that $\u^\star$ has the same sign as $\x$.
From now on, we can therefore assume that $\x \in \RR_+^n$,
which implies that $\u \in \RR^n_+$.

Since $\u \in \RR^n_+$ and $\phi$ is increasing on $\RR_+$, we have
$u_{\sigma_1} \ge \dots \ge u_{\sigma_n} 
\Rightarrow 
\phi(u_{\sigma_1}) \ge \dots \ge \phi(u_{\sigma_n})$.
We know that for all $\u \in \RR^n$ and $\w \in \RR^n$, we have
$f(\varphi(u), \w) 
= \langle \varphi(\u)_\sigma, \w \rangle 
= \langle \varphi(\u_\sigma), \w \rangle$,
where $\sigma$ is the permutation sorting $\u$ in descending order.
From now on, let us fix $\sigma$ to the permutation sorting $\u^\star$.
We will show in the sequel that this is the same permutation as the one
sorting $\x$. We then have
\begin{equation}
f_{\varphi,R}(\x, \w) 
= \min_{\u \in \RR^n} R^*(\x-\u) + \langle \varphi(\u_\sigma), \w \rangle.
\end{equation}
Using the change of variable
$\v = \u_\sigma \Leftrightarrow \v_{\sigma^{-1}} = \u$,
we obtain
\begin{equation}
f_{\varphi,R}(\x, \w) =
\max_{v_1 \ge \dots \ge v_n}
R^*(\x_\sigma - \v) + \langle \varphi(\v), \w \rangle
\end{equation}
where we used that if $R(\y) = \sum_{i=1}^n r(y_i)$, then
\begin{equation}
R^*(\x - \u) = R^*(\x - \v_{\sigma^{-1}}) = R^*(\x_\sigma - \v).
\end{equation}
Let $\s \coloneqq \x_\sigma$. It remains to show that $s_1 \ge \dots \ge s_n$,
i.e., that $\s$ and $\v^\star$ are both in descending order. 
Suppose $s_j > s_i$ for some $i < j$.
Let $\s'$ be a copy of $\s$ with $s_i$ and $s_j$ swapped.
Since $R^*$ is convex,
by \citep[Lemma 4]{blondel2020fast},
\begin{equation}
R^*(\s - \v^\star) - R^*(\s' - \v^\star)
= r^*(s_i - v_i^\star) + r^*(s_j - v_j^\star)
- r^*(s_j - v_i^\star) - r^*(s_i - v_j^\star)
\ge 0,
\end{equation}
which contradicts the assumption that $\v^\star$ 
and the corresponding $\sigma$ are optimal.



\subsection{Subproblem derivation}
\label{proof:example_isotonic}

\paragraph{Case $\phi(x) = \frac12 x^2$ and $R(\x) = \frac\lambda2\|\x\|^2$.}One has $h_i(\gamma) = \frac12 (w_i \gamma^2 + \frac1\lambda(s_i - \gamma)^2)$ so that $\frac {d h_i}{d \gamma} = (w_i + \frac1\lambda) \gamma - \frac1\lambda s_i.$
Therefore, 
$$
\frac {d \sum_{i\in B}h_i(\gamma)}{d \gamma} = \gamma \sum_{i\in B} (w_i + \frac1\lambda) - \frac1\lambda \sum_{i\in B} s_i.
$$
Since in addition $\sum_{i\in B}h_i$ is convex we obtain that its minimum is given by canceling the derivative, hence 
$$
\gamma_B^\star = \frac{ \sum_{i\in B} s_i}{\sum_{i\in B} (\lambda w_i + 1)}.
$$

\paragraph{Remark.}

This result shows that when $\lambda$ is small enough, one can control the approximation error induced by our proposed operator in comparison to the hard operator. For simplicity, let us focus on the case where there are no ties: $\forall i \neq j$, $x_i \neq x_j$. This implies that $s_1 > s_2 > \cdots > s_n$. In this case, for $\lambda$ small enough, we get $$\gamma_{\{1\}}^\star > \gamma_{\{2\}}^\star > \cdots > \gamma_{\{n\}}^\star$$ so that  the optimal partition in PAV's algorithm is given by taking $B_i = \{ i\}$. Therefore, $v_i^\star = \frac{s_i}{\lambda w_i + 1}.$ 
One then has 
$$
\u^\star = \mathrm{sign}(\x) \circ \v^\star_{\sigma^{-1}} = \frac{\x}{\lambda \w_{\sigma^{-1}} + 1}.
$$
Plugging it into $\y^\star$ given by Proposition \ref{prop:relaxed_operator} gives 
$$
\y^\star = \frac{\w_{\sigma^{-1}} \circ \x }{\lambda \w_{\sigma^{-1}} + 1}.
$$
Since the hard operator is given by $\w_{\sigma^{-1}} \circ \x $, the approximation error $e_{\lambda}(\x, \w)$ in infinite norm is then simply bounded by
$$
e_{\lambda}(\x, \w) := \|   \w_{\sigma^{-1}} \circ \x (\frac{1}{\lambda \w_{\sigma^{-1}} + 1} -1) \|_{\infty} \leq \|\w\|_{\infty}\|\x\|_{\infty}\|\frac{\lambda \w}{\lambda \w + 1}\|_{\infty} .
$$
In the $\topkmag$ case, $w = \ones_k$, so that $e_{\lambda}(\x, \w) \leq \lambda \|\x\|_{\infty}$.
\paragraph{Case $\phi(x) = x$ and $R(\x) = \frac\lambda p\|\x\|^p$ with $p = \frac43$.} One has $h_i(\gamma) = w_i \gamma + \frac1{4 \lambda^3}(s_i - \gamma)^4$ so that $\frac {d h_i}{d \gamma} = w_i + \frac{1}{\lambda^3}(\gamma - s_i)^3.$
Therefore, 
$$
\frac {d \sum_{i\in B}h_i(\gamma)}{d \gamma} =  \frac{1}{\lambda^3} \sum_{i\in B}(\gamma - s_i)^3  + \sum_{i\in B} w_i.
$$
Since in addition $\sum_{i\in B}h_i$ is convex we obtain that its minimum is given by canceling the derivative, and hence by solving the third-order polynomial equation 
$$
 \frac{1}{\lambda^3} \sum_{i\in B}(\gamma - s_i)^3  + \sum_{i\in B} w_i = 0.
$$
In practice, we solve this equation using the root solver from the $\textit{numpy}$ library. Note that taking $p=\frac43$ leads to an easier subproblem than $p=\frac32$, hence our choice for $p$.

\paragraph{Case $\phi(x) = \frac12 x^2$ and $R(\x) = \frac\lambda p\|\x\|^p$ with $p = \frac43$.} The derivation is very similar to the previous case. Indeed, one has $h_i(\gamma) = \frac12 w_i \gamma^2 + \frac1{4 \lambda^3}(s_i - \gamma)^4$ so that $\frac {d h_i}{d \gamma} = w_i \gamma + \frac{1}{\lambda^3}(\gamma - s_i)^3.$
Therefore, 
$$
\frac {d \sum_{i\in B}h_i(\gamma)}{d \gamma} =  \frac{1}{\lambda^3} \sum_{i\in B}(\gamma - s_i)^3  + \gamma \sum_{i\in B} w_i.
$$

\subsection{Differentiation - Proof of Proposition \ref{prop:jacobian}}
\label{proof:jacobian}

\paragraph{Case $\phi(x) = x$.} One has $h_i(\gamma) = \frac1q |s_i - \gamma|^q + w_i v_i.$ For any optimal $B$ in PAV one has the optimality condition 
$$
\sum_{i \in B} (\mathrm{sign}(\gamma^\star_B - s_i)|\gamma^\star_B -s_i|^{q-1} + w_i) = 0.
$$
Using the implicit function theorem gives that $\gamma^\star_B$ is differentiable, and differentiating with respect to any $s_i$ for $i \in B$ leads to 
$$
\frac{\partial \gamma^\star_B}{\partial s_i} = \frac{|\gamma^\star_B - s_i|^{q-2}}{\sum_{j\in B} |\gamma^\star_B - s_j|^{q-2}}.
$$
\paragraph{Case $\phi(x) = \frac12 x^2$.} Similar calculations lead to 
$$
\frac{\partial \gamma^\star_B}{\partial s_i} = \frac{(q-1)|\gamma^\star_B - s_i|^{q-2}}{\sum_{j\in B} (q-1)|\gamma^\star_B - s_j|^{q-2} + w_j}.
$$

\subsection{Dual of isotonic optimization}
\label{proof:isotonic_reg_dual}

\begin{equation}
\begin{aligned}
\min_{v_1 \ge \dots \ge v_n} \sum_{i=1}^n h_i(v_i)
&= \min_{\v \in \RR^n} 
\max_{\alphav \in \RR^{n-1}_+} \sum_{i=1}^n h_i(v_i) - \alpha_i (v_i - v_{i+1}) \\
&= \min_{\v \in \RR^n} 
\max_{\alphav \in \RR^{n-1}_+} \sum_{i=1}^n h_i(v_i) - v_i (\alpha_i - \alpha_{i-1}) \\
&= \max_{\alphav \in \RR^{n-1}_+} -\left[ 
\sum_{i=1}^n h^*_i(\alpha_i - \alpha_{i-1})
\right]
\end{aligned}
\end{equation}
where $\alpha_0 \coloneqq 0$ and $\alpha_n \coloneqq 0$ are constants (i.e., not optimized).
An optimal solution $\v^\star$ is recovered from $\alphav^\star$ by 
$v^\star_i = (h_i^*)'(\alpha_i^\star - \alpha_{i-1}^\star)$.
Since $\alphav$ is only constrained to be non-negative, we can solve the dual by coordinate ascent.
The subproblem associated with $\alpha_i$, for $i \in \{1, \dots, n-1\}$, is 
\begin{equation}
\max_{\alpha_i \in \RR_+} -h_i^*(\alpha_i - \alpha_{i-1}) - h_{i+1}^*(\alpha_{i+1} - \alpha_i).
\end{equation}
The subproblem is a simple univariate problem with non-negative constraint. Let us define $\alpha_i'$ as the solution
of $(h_i^*)'(\alpha_i' - \alpha_{i-1}) -(h_{i+1}^*)'(\alpha_{i+1} - \alpha_i') = 0$. The solution is then $\alpha_i^\star = [\alpha_i']_+$.

In practice, we can alternate between updating $\alpha_1, \alpha_3, \dots$ in parallel and 
$\alpha_2, \alpha_4, \dots$ in parallel.
The dual variables with odd coordinates correspond to the set
$C_1 = \{\v \in \RR^n \colon v_1 \ge v_2, v_3 \ge v_4, \dots \}$
and the dual variables with even coordinates correspond to the set
$C_2 = \{\v \in \RR^n \colon v_2 \ge v_3, v_4 \ge v_5, \dots \}$.
Coordinate ascent converges to an optimal dual solution,
assuming each $h_i^*$ is differentiable, which is equivalent to each $h_i$ being strictly convex.

Note that the subproblem can be rewritten in primal space as
\begin{equation}
\begin{aligned}
&\max_{\alpha_i \in \RR_+} -h_i^*(\alpha_i - \alpha_{i-1}) - h_{i+1}^*(\alpha_{i+1} - \alpha_i) \\
&=\max_{\alpha_i \in \RR_+} 
-\left[ \max_{v_i} (\alpha_i - \alpha_{i-1}) v_i - h_i(v_i) \right]
-\left[ \max_{v_{i+1}} (\alpha_{i+1} - \alpha_i) v_{i+1} - h_{i+1}(v_{i+1}) \right] \\
&= \min_{v_i,v_{i+1}} \alpha_{i-1} v_i + h_i(v_i) - \alpha_{i+1} v_{i+1} + h_{i+1}(v_{i+1}) + \max_{\alpha_i \in \RR_+} \alpha_i (v_{i+1} - v_i) \\
&= \min_{v_i \ge v_{i+1}} h_i(v_i) + h_{i+1}(v_{i+1}) + \alpha_{i-1} v_i - \alpha_{i+1} v_{i+1}.
\end{aligned}
\end{equation}
In fact, in the Euclidean case, it is known that Dykstra's algorithm in the primal and block coordinate ascent in the dual are equivalent \citep{tibshirani_2017}.
Therefore, block coordinate ascent can be seen as an elegant way to generalize Dykstra's algorithm to the non-Euclidean case.

\section{Additional material}

\subsection{k-support negentropies}
\label{app:k_support_negent}

When $\phi(x) = e^{x - 1}$ and $\w = \ones_k$, we obtain 
\begin{equation}
f_\varphi^*(\y, \w) = \min_{\z \in [0, 1]^n} \sum_{i=1}^n y_i \log(\frac{y_i}{z_i})
\quad \text{s.t.} \quad
\langle \z, \ones \rangle = k.
\end{equation}
We call it a $k$-support negative entropy.

\begin{figure}[h]
\begin{center}
\includegraphics[scale=0.7]{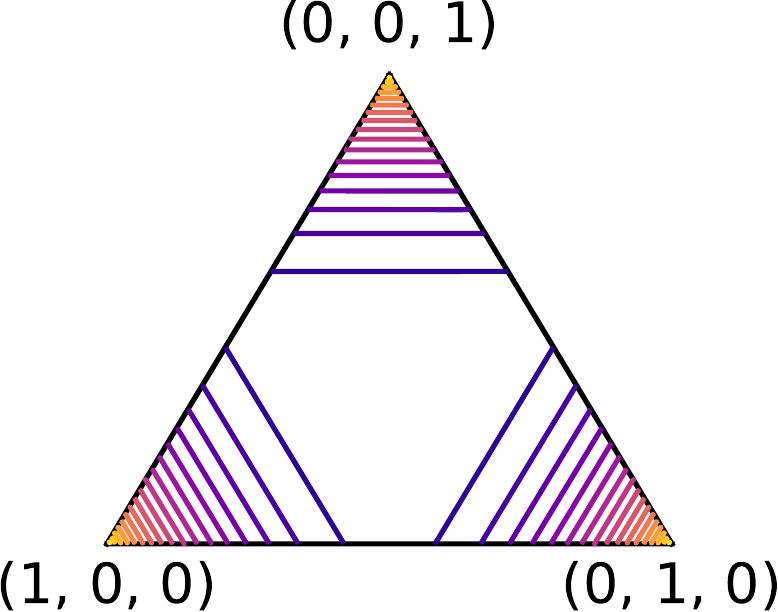}
  \end{center}
\caption{\textbf{Contours of the k-support entropy} on the simplex for $n=3$ and
$k=2$. Lighter colors indicate lower values.}
\label{fig:ent}
\end{figure}
\subsection{PAV algorithm}\label{app:pav}
We present the pseudo code for PAV, adapted from \citet{lim2016efficient}. Recall that we define 
\begin{equation}
\gamma^\star_B = \argmin_{\gamma \in \RR}\sum_{i \in B}h_i(\gamma).
\end{equation} 
\begin{algorithm}
\caption{Pool Adjacent Violators (PAV)}
   \label{alg:example}
\begin{algorithmic}
   \STATE {\bfseries Input:} Convex functions $\{h_i$ : $\RR \to \RR\}_{i\in[n]}$\\
   Initalize partitions $P \leftarrow  \{\{i\} | i \in[n]\}$\\
   Initialize $v_i \leftarrow \gamma_{\{i\}}^\star$ for all $i \in [n]$
   \WHILE {there exists $i$ such that $v_i  < {v_{i+1}}$}
   \STATE Find $B_{r_i}$ and $B_{r_{i+1}}$ in $P$ such that $i \in B_{r_i}$ and $i+1 \in B_{r_{i+1}}$ \\
   Remove $B_{r_i}$ and $B_{r_{i+1}}$ from $P$ \\
   Add $B_{r_i} \cup B_{r_{i+1}}$ to $P$ \\
   Compute $\gamma_{B_{r_i} \cup B_{r_{i+1}}}^\star$ by solving \eqref{eq:pav_subproblem} \\
   Assign $v_r \leftarrow \gamma_{B_{r_i} \cup B_{r_{i+1}}}^\star$ for all $r \in B_{r_i} \cup B_{r_{i+1}}$ 
   \ENDWHILE
   \STATE {\textbf{return} $v$}
\end{algorithmic}
\end{algorithm}

\section{Experimental details}\label{app:exp_details}

\subsection{Weight pruning in neural networks} 
For our experiment on the MNIST dataset, we train the MLP using SGD with a batch size of $128$ and a constant learning rate of $10^{-2}$. 
We trained the model for $30$ epochs.
In terms of hardware, we use a single GPU. 

\subsection{Smooth top-k loss} 
For our experiment on the CIFAR-100 dataset, we train the ViT-B/16 using SGD with a momentum of 0.9 and with a batch size of $512$. 

For the cross-entropy loss, we follow the training procedure of \citet{dosovitskiy2020image}: warmup phase until the learning rate reaches $3 \times 10^{-3}$. The model is trained for $100$ steps using a cosine learning rate scheduler. This choice of learning rate gave the best performance for this number of training steps.

For our top-$3$ losses: warmup phase until the learning rate reaches $5 \times 10^{-3}$. The model is trained for $100$ steps using a cosine learning rate scheduler.

In terms of hardware, we use $8$ TPUs.

\subsection{Sparse MoEs}

We train the V-MoE S/32 model \citep{riquelme2021scaling} on the JFT-300M dataset \citep{sun2017revisiting}. JFT is a multilabel dataset, and thus accuracy is not an appropriate metric since each image may have multiple labels. Therefore, we measure the quality of the models using the commonly-used \emph{precision-at-1} metric \citep{jarvelin2017ir}.
The training procedure is analogous to the one described in \citet{riquelme2021scaling}, except that we replace the routing algorithm. In particular, we use the Expert Choice Routing algorithm described in \citet{zhou2022mixture} as our baseline, and replace the non-differentiable top-$k$ operation used there with our differentiable approach (we perform 10 iterations of Dykstra's algorithm).

We use exactly the same hyperparameters as described in \citet{riquelme2021scaling}, except for the fact that Expert Choice Routing does not require any auxiliary loss. Specifically, we train for 7 epochs using a batch size of 4\,096. We use the Adam optimizer ($\beta_1$ = 0.9, $\beta_2$ = 0.999), with a peak learning rate of $10^{-3}$, warmed up for 10\,000 steps and followed by linear decay. We use mild data augmentations (random cropping and horizontal flipping) and weight decay of $10^{-1}$ in all parameters as means of regularization.  We trained both models on TPUv2-128 devices.

\end{document}